\documentclass[a4paper,twoside]{article}

\usepackage{epsfig}
\usepackage{subcaption}
\usepackage{calc}
\usepackage{amssymb}
\usepackage{amstext}
\usepackage{amsmath}
\usepackage{amsthm}
\usepackage{multicol}
\usepackage{pslatex}
\usepackage{apalike}
\usepackage{algorithm2e}
\usepackage[bottom]{footmisc}
\usepackage{SCITEPRESS}     % Please add other packages that you may need BEFORE the SCITEPRESS.sty package.
\usepackage{tabularx}
\usepackage{cite}
\usepackage{comment}
\usepackage{orcidlink}
\usepackage{verbatim}
\usepackage{textcomp}
\usepackage{blindtext}
\usepackage{nicefrac}
\usepackage{stfloats}

\def\BibTeX{{\rm B\kern-.05em{\sc i\kern-.025em b}\kern-.08em
T\kern-.1667em\lower.7ex\hbox{E}\kern-.125emX}}

\begin{document}

\title{Coconut Palm Tree Counting on Drone Images with Deep Object Detection and Synthetic Training Data}

\author{\authorname{Tobias Rohe\sup{1}\orcidlink{0009-0003-3283-0586}, Barbara Böhm\sup{1}, Michael Kölle\sup{1}\orcidlink{0000-0002-8472-9944}, Jonas Stein\sup{1,2}\orcidlink{0000-0001-5727-9151}, Robert Müller\sup{1}, and Claudia Linnhoff-Popien\sup{1}\orcidlink{0000-0001-6284-9286}}
\affiliation{\sup{1}Mobile and Distributed Systems Group, LMU Munich, Germany}
\affiliation{\sup{2}Aqarios GmbH, Germany}
\email{tobias.rohe@ifi.lmu.de}
}


\abstract{
Drones have revolutionized various domains, including agriculture. Recent advances in deep learning have propelled among other things object detection in computer vision. This study utilized YOLO, a real-time object detector, to identify and count coconut palm trees in Ghanaian farm drone footage. The farm presented has lost track of its trees due to different planting phases. While manual counting would be very tedious and error-prone, accurately determining the number of trees is crucial for efficient planning and management of agricultural processes, especially for optimizing yields and predicting production. We assessed YOLO for palm detection within a semi-automated framework, evaluated accuracy augmentations, and pondered its potential for farmers. Data was captured in September 2022 via drones. To optimize YOLO with scarce data, synthetic images were created for model training and validation. The YOLOv7 model, pretrained on the COCO dataset (excluding coconut palms), was adapted using tailored data. Trees from footage were repositioned on synthetic images, with testing on distinct authentic images. In our experiments, we adjusted hyperparameters, improving YOLO's mean average precision (mAP). We also tested various altitudes to determine the best drone height. From an initial mAP@.5 of $0.65$, we achieved 0.88, highlighting the value of synthetic images in agricultural scenarios.
}

\onecolumn \maketitle \normalsize \setcounter{footnote}{0} \vfill

\section{INTRODUCTION} \label{sec:introduction}

Coconut farming, pivotal to West African economies, offers both a sustainable livelihood and essential contributions to regional food systems. However, challenges in monitoring the growth and count of coconut palm trees, increased by varied planting phases and environmental conditions, pose significant operational hindrances for larger farms. This paper presents an innovative approach to address these challenges by leveraging deep learning to detect and count coconut palm trees using drone imagery, specifically focusing on a farming project in the eastern region of Ghana.

Initiated in August 2021, the farming project aimed to cultivate approximately 2,500 coconut palm trees alongside other crops. The overarching goal was not only to foster a sustainable family-run coconut farming business but also to bolster employment opportunities and social benefits for local communities, thereby intertwining traditional farming practices with modern techniques. However, as the farm expanded, maintaining an accurate count of the trees became a daunting task, with manual surveys proving both time-consuming and error-prone.

Addressing this, we explored the application of computer vision, a subfield of computer science focusing on replicating human vision system capabilities, to detect and enumerate the coconut palm trees. While classical object detection methods relied on handcrafted features, recent advancements in machine learning and deep learning have revolutionized this space, with techniques such as YOLO (You Only Look Once) offering enhanced accuracy and efficiency.

This study delves into the application of the YOLOv7 framework, released in 2022, to our specific use case. Beyond mere counting, future applications of this methodology might extend to discerning the health of plants, thereby offering comprehensive farm management solutions. By focusing on a real-world problem, this work aims to bridge the gap between advanced technical solutions and practical agricultural challenges, setting the stage for more integrative, technology-driven farming practices in the future. In summary, our contributions are:

\begin{enumerate}
    \item We introduce a novel real world application: Finetuning a deep object detector to count coconut palm trees.
    \item We show that performance can substantially be increased by not only considering coconut palm trees during training but also other plants. This allows the object detector to better differentiate between the plants it sees.
    \item We show that it suffices to train on synthetically generated images and thereby eliminate the need to manually label the images.
\end{enumerate}
\section{BACKGROUND}\label{sec:Fundamentals and State of the Art in Object Detection}

This section delves into the foundational concepts and the current state of object detection. We start with Convolutional neural networks. Subsequently, our focus shifts to object detection, a critical component to address our use case. In this context, we enumerate the four scenarios encountered during object detection, especially when positioning bounding boxes. We also elucidate the concept of Intersection-Over-Union. Building on this, we discuss key performance metrics in object detection, namely precision, recall, and mean average precision. Concluding this section, we present the YOLO architecture, emphasizing the advancements in YOLOv7—a state-of-the-art object detector. The experiments we discuss in Section 4 predominantly leverage this technology.

\subsection{Convolutional Neural Networks}

Convolutional Neural Networks (CNNs) have emerged as a pivotal technology in the domain of computer vision. Originating from the larger family of Deep Neural Networks (DNNs), CNNs are specifically tailored for image data, making them adept at tasks such as face recognition, image classification, and object detection.

A standard CNN architecture comprises several layers. The inaugural layer, the convolutional layer, is pivotal for feature extraction. By utilizing filter matrices, or kernels, this layer captures intricate patterns such as colours and edges from an image. The unique property of these filters is their translational invariance, allowing objects to be recognized irrespective of their spatial positioning in an image.

Following the convolutional layer is the pooling layer, designed for dimensionality reduction. This dimension reduction not only decreases the computational burden but also helps in extracting dominant features. Two common pooling methods exist: max pooling and average pooling, which respectively capture the maximum and average values from a designated window in the input.

In the deeper sections of the network, fully-connected layers serve the crucial role of integrating features from previous layers and mapping them to the desired output. These layers essentially form the decision-making component of the CNN.

Training a CNN involves defining its architecture and then optimising it over several iterations, known as epochs. Throughout this training phase, the model's internal parameters get refined to enhance prediction accuracy. This training is supervised, requiring labelled datasets to guide the iterative minimisation of prediction errors. Within the realm of CNNs, renowned architectures include AlexNet, GoogLeNet, and VGGNet.

Critical challenges in training CNNs encompass phenomena like underfitting and overfitting. Addressing these challenges, often by tuning hyperparameters, ensures that the trained model is both robust and accurate in its predictions~\cite{SEV21}.

\subsection{Object Detection}

Computer vision, pivotal in numerous applications, encompasses tasks such as image classification, segmentation, and object detection. Object detection marries object recognition — identifying and classifying objects within media — with object localization, which encapsulates these identified objects within bounding boxes~\cite{KHA20}. A key metric here is the Intersection-over-Union (IoU), which quantifies the overlap between the predicted bounding box and the ground truth — the actual annotated bounding box. It calculates the ratio of their intersection to their union. IoU values range between 0 and 1: values close to 0 imply minimal overlap and those nearing 1 signify accurate predictions~\cite{ANW22}.

\subsection{YOLO - You only look once}

This section delves into modern object detectors, focusing on YOLO (You Only Look Once). The field of object detection began with Region-based Convolutional Network (R-CNN) in 2014. It proposed regions and fed them into a classifier, a methodology advanced by its successors: Fast R-CNN, Faster R-CNN, and Mask R-CNN. These methods identified potential regions of interest and subsequently classified them for detection~\cite{Szeliski:2022:CVA}.

YOLO, introduced in 2016, diverges from this two-step approach. It performs real-time object detection by predicting bounding boxes and class probabilities directly from images in a single evaluation, enhancing speed and requiring fewer resources~\cite{redmon2016you}. The model encompasses $24$ convolutional layers followed by $2$ fully connected layers, leveraging a rectified linear activation function in all but the final layer~\cite{redmon2016you}. YOLO's architecture breaks the input image into an $S \times S$ grid, making predictions encoded as an $S \times S \times (B*5+C)$ tensor~\cite{redmon2016you}.

Our experiments in section $4$ employ YOLOv7, released in 2022. This variant, pretrained on the MS Coco Dataset with 80 object categories, lacks a category for coconut palm trees, necessitating custom data for our use case: counting coconut palm trees in drone images. YOLOv7 boasts innovations like the extended efficient layer aggregation network (E-ELAN) and a trainable bag of freebies, enhancing speed and accuracy without elevating costs~\cite{RHA22}. The architecture, with around 37 million parameters, features a backbone established in the first 50 layers.

\section{METHOD}\label{sec:Concept and Workflow}

In this section our concept and the derived workflow is described. Both are geared towards answering the overarching question of the use case. The concept and workflow was chosen as follows and reflects the typical phases of a deep learning project.

\subsection{Data Acquisition}

In September 2022, aerial drone images were captured in Ghana at heights of 10m, 25m, 45m, 70m, and 85m. These 73 images, captured using a DJI Mini 2 drone camera, have a resolution of 4000x2250 pixels. Given the common data scarcity challenge in deep learning and the guideline suggesting roughly 5000 labeled examples per category for adequate performance~\cite{Goodfellow:2016:DL}, we generated additional data synthetically. We based this on 13 selected drone images, keeping the remaining images for model testing. 

\subsection{Data Preparation and Generation}

Data preparation follows a cross-validation approach, partitioning the data into training, validation, and test sets~\cite{Prince:2023:UDL}. Due to data limitations, we generated synthetic images to train YOLOv7. Using GIMP, specific plants were extracted from the raw images, while Dall-E produced the backgrounds (BG). A custom Python generator then assembled these components into synthetic images by:

\begin{itemize}
   \item Randomly selecting BGs and plants
   \item Applying random plant rotation, size, and flip
   \item Positioning plants on the BG without overlaps
   \item Adjusting the plant count per BG based on configuration
\end{itemize}

This generator also produces YOLO-formatted label files, essential for training and validation. We uploaded the synthetic and test images to Google Drive, organizing them into specific directories for compatibility with YOLOv7.

\subsection{Metric}
The precision-recall curve graphically illustrates the trade-off between the two metrics for different confidence thresholds. As a reminder these metrics are utilised to evaluate the performance of deep learning models. The curve is typically depicted in a coordinate system where the x-axis is the recall and the y-axis is the precision and both values are always between 0 and 1. The graph shows how both metrics change as the threshold is adjusted. A higher curve generally points out better model performance. Additionally, the area underneath the curve is known as the average precision (AP)~\cite{Glassner:2021:DL}. 

AP is an essential performance indicator for one object class, whereas the mean average precision (mAP) is used when detecting multiple object categories in an image. It is calculated by the sub metrics confusion matrix, IoU, recall and precision. ``For each class $k$, we calculate the mAP across different IoU thresholds, and the final metric mAP across test data is calculated by taking an average of all mAP values per class.''~\cite{SHA22}. Higher values show better performance in object detection tasks. The $mAP$ is defined as:

$$ mAP = \frac{1}{n}\sum_{k=1}^{k=n} AP_{k}$$

\noindent where $AP_{k}$ denotes the average precision of class $k$ and $n$ is the number of classes.
In subsequent sections we use mAP@.5 to compare the outcome of the experiments, which means a single IoU-threshold of $0.5$ was applied. The mAP@.5 value seemed appropriate for the use case, since counting is more essential than precise localisation.   

\subsection{Training, Validation and Test}

Model training, validation, and testing occurred on Google Colab. We installed YOLOv7 and its dependencies on the Colab virtual machine and utilized pre-trained COCO dataset weights. After adjusting the necessary configuration files, the model underwent training and subsequent testing for mAP results. TensorBoard helped in monitoring the training process.

\section{Results}\label{sec:Results}

This study conducted a series of experiments aiming to enhance the mAP at an IoU threshold of $0.5$ by refining the workflow parameters. We established a baseline using coconut palm trees as the primary object class.

\subsection{Baseline}

Establishing a baseline is pivotal for contextualizing model performance and offering a point of reference for subsequent experiments\cite{NAI22}. Given Ghana's reddish laterite soil, DALL-E simulated these background colors. Focusing on the objective of counting coconut palm trees, we initially considered this as the sole object class. We generated 300 synthetic images for training, featuring between $15$ to $25$ trees each, totaling approximately $6000$ trees. After $40$ epochs of training using pre-trained YOLOv7 weights, we achieved a mAP@.5 value of $0.65$, detailed in the subsequent section.

\subsection{Varying Background Colors}

Healthy coconut trees possess vibrant green leaves with patches of brown, with color variations influenced by health, age, and environment. Notably, West Africa's laterite soil presents a reddish hue, while surrounding vegetation remains chiefly green.

To ascertain the optimal background texture or dominant color yielding the highest results with trained YOLO models, backgrounds for training and validation were crafted using stable diffusion, utilizing DALL-E, an AI that translates natural language descriptions into images. As illustrated in Table \ref{tab:Background Color}, the prompts were: dense green vegetation, red soil (offering enhanced contrast with the trees), and a combination of the two.

\begin{figure}[hpbt]
 \centering
  %%----start of first subfigure----
  \subfloat[Green Background]{
   \label{fig:green:a} %% label for first subfigure
   \includegraphics[width=0.46\linewidth]{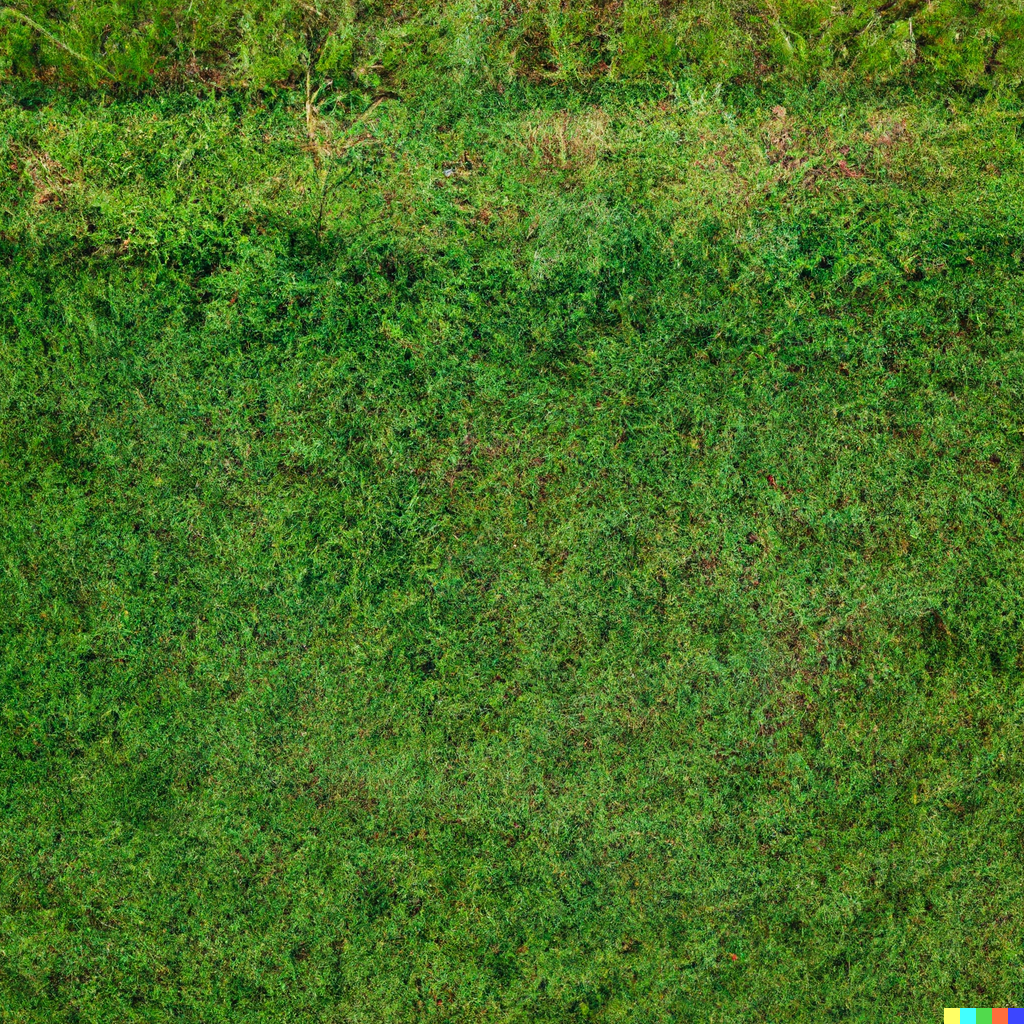}}
  \hspace{0.01\linewidth}
  %%----start of second subfigure----
  \subfloat[Red Background]{
   \label{fig:red:b} %% label for second subfigure
   \includegraphics[width=0.46\linewidth]{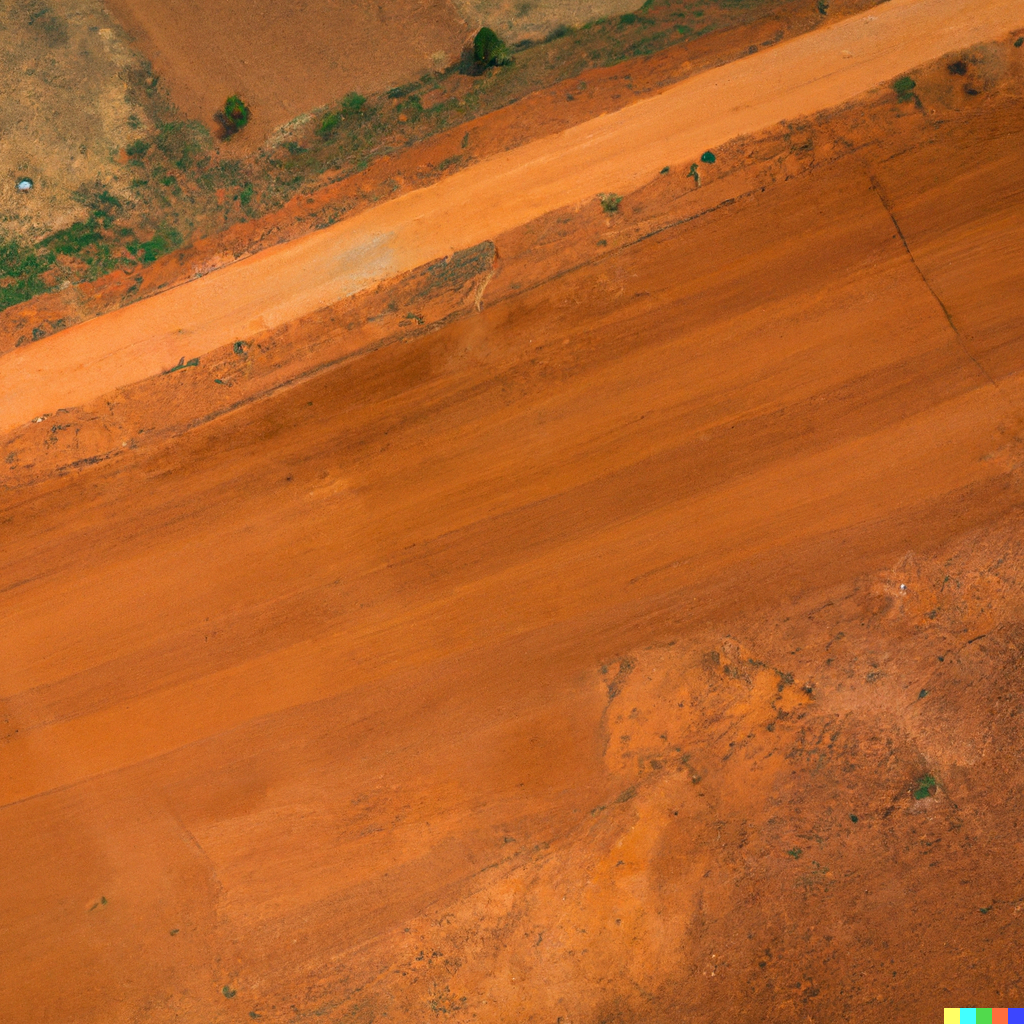}}\\[0pt] % horizontal break
 \caption[Example of Stable Diffusion output]{Example of Stable Diffusion output}
 \label{fig:Exp1} %% label for entire figure
\end{figure}

Figure \ref{fig:Exp1} displays samples of green and red backgrounds. Each experiment iteration followed a similar procedure. The column "Size of BG pool" specifies the backgrounds available for test (T) and validation (V) datasets. Palm trees, rotated and placed at random, ranged between $15$ to $25$ per image. Models were trained over $40$ epochs with a batch size of $16$.

%\ref{tab:Background Color}.

\begin{table*}[htb]
{\small
\begin{center}
\begin{tabularx}{\linewidth}{|l|X|l|}
\hline
\textbf{BG Color} & \textbf{Prompt} & \textbf{Size of BG Pool}\\
\hline
green & "vertical aerial drone picture of ghana farm grass" & T:35 V:21\\
red & "vertical aerial drone pictures of ghana soil" & T:35 V:21\\
mixed & equal composition of a and b & T:35 V:21\\
\hline
\end{tabularx}
\end{center}
}
\caption[Prompts to create background with DALL-E]{Prompts to create background with DALL-E \label{tab:Background Color}}
\end{table*}

%\ref{fig:Exp1}

To account for result variability, models were thrice tested on data taken $25$ meters above ground. Tests utilized the optimal model from prior training in terms of mAP on the validation set~\cite{SKE22} with an object confidence threshold set at $0.6$.

\begin{figure}[hpbt]
 \centering
  %%----start of first subfigure----
  %\subfloat[different backgrounds colors]{
   \label{fig:exp1:a} %% label for first subfigure
   \includegraphics[width=\linewidth]{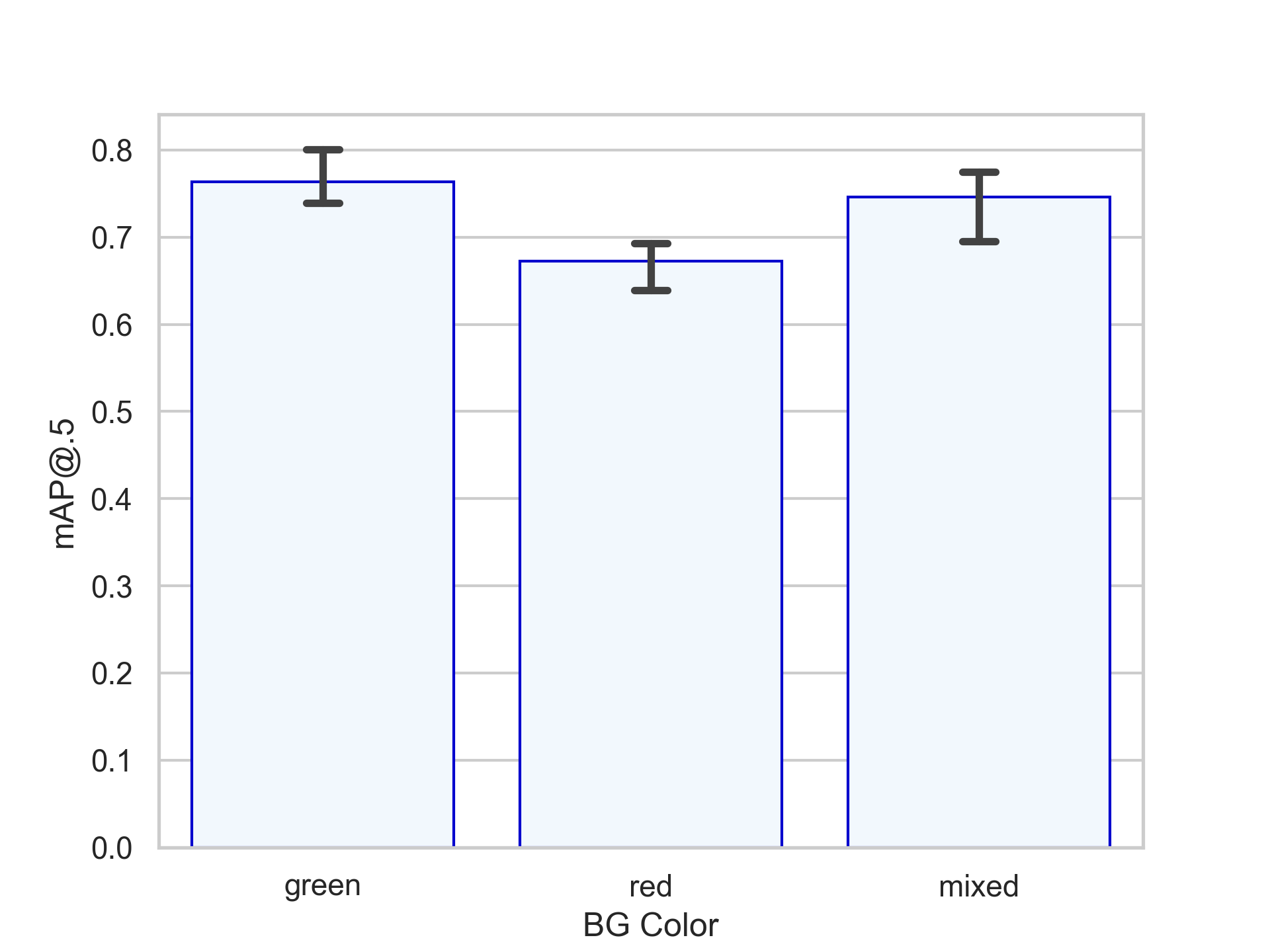}
 \caption[mAP@.5 values for different BG]{mAP@.5 values for different BG}
 \label{fig:Exp1.1} %% label for entire figure
\end{figure}

Figure       \ref{fig:Exp1.1}'s bar plot delineates the mAP@0.5 values, revealing that the green background, akin to actual drone imagery grounds, outperformed others and thus was selected for subsequent experiments.

\begin{figure}[hpbt]
 \centering
  %%----start of first subfigure----
  %\subfloat[detection result of green bg]{
   \label{fig:expgreenbg:a} %% label for first subfigure
   \includegraphics[width=0.90\linewidth]{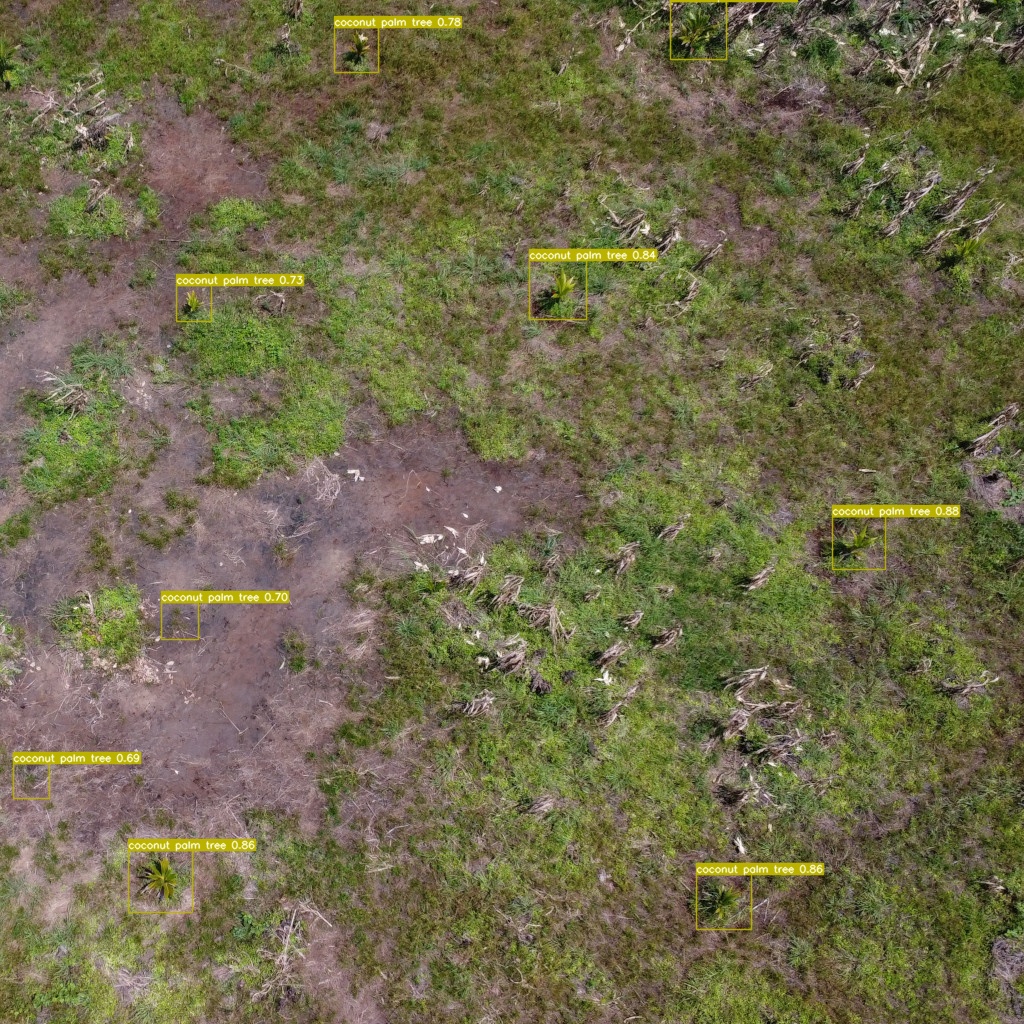}
  \hspace{0.01\linewidth}
 \caption[Output of detect command]{Output of detect command}
 \label{fig:Exp2} %% label for entire figure
\end{figure}

Figure~\ref{fig:Exp2} showcases the detect command results for the green background, indicating fewer false positives/negatives than its counterparts. Raising the confidence threshold to $0.71$ further refined results. A potential inference from Figure \ref{fig:Exp1.1} is that the model's capacity to discern palm trees improves substantially when the stark contrast of green trees against a red backdrop isn't the dominant visual cue, but rather the model has to learn other features for detection.

%\ref{fig:Exp1.1}

%\ref{fig:Exp2}

\subsection{Multiple Object Classes}

\begin{figure*}[hpbt]
 \centering
  %%----start of first subfigure----
  \subfloat[palm tree]{
   \label{fig:palmtree:a} %% label for first subfigure
   \includegraphics[width=0.23\linewidth]{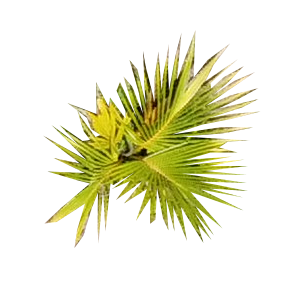}}
  \hfill
  %%----start of second subfigure----
  \subfloat[okra]{
   \label{fig:okra:b} %% label for second subfigure
   \includegraphics[width=0.23\linewidth]{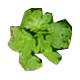}} % horizontal break
   \hfill
  %%----start of third subfigure----
  \subfloat[tree trunk]{
   \label{fig:treetrunk:c} %% label for third subfigure
   \includegraphics[width=0.23\linewidth]{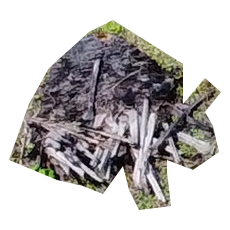}}
  \hfill
  %%----start of fourth subfigure----
  \subfloat[patch of grass]{
   \label{fig:grass:d} %% label for fourth subfigure
   \includegraphics[width=0.23\linewidth]{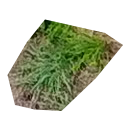}}
 \caption{Plant cut outs used as overlay}
 \label{fig:Exp3} %% label for entire figure
\end{figure*}

While the initial approach focused solely on identifying coconut palm trees, certain plants like okra and weeds were mistakenly identified as coconut palm trees. This experiment aimed to determine if incorporating these misidentified objects as separate classes would enhance the mAP@0.5 accuracy for the primary object – the coconut palm trees. Table \ref{tab:Object Class} lists the variants, Figure \ref{fig:Exp3} their visual representations. 

Okra, a staple in Ghanaian cuisine, can grow up to 2 metres tall and has a notable presence in Ghana, ranking it among the top ten global okra producers. Its distinct shape, as well as that of tree trunks, eases the generation of training data. However, tree trunks and patches of grass, which had prior misidentified, were isolated and labelled. This necessitated extending the Python generator to accommodate multiple object classes, as reflected in Table \ref{tab:Object Class} with an aggregate of $2,911$ plants.

%\ref{fig:Exp3}

\begin{figure}
    \centering
    \includegraphics[width=\linewidth]{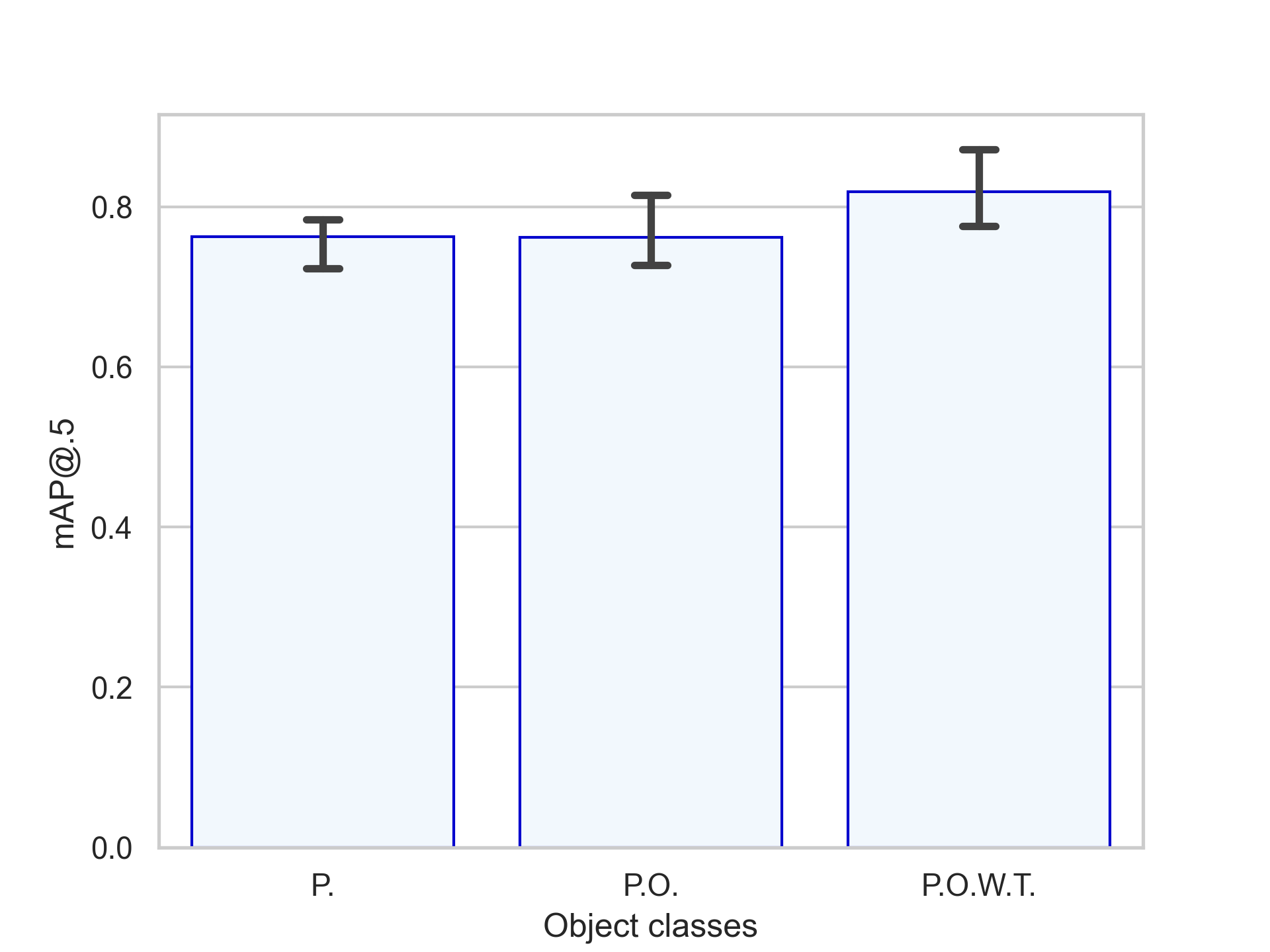}
    \caption{Test results with one, two and four classes}
    \label{fig:exp2:a}
\end{figure}

Maintaining the green background, the workflow remained consistent across trials. Results in Figure~\ref{fig:exp2:a} suggest the four-class model offers superior mAP@.5 values. Furthermore, it demonstrates effective okra recognition alongside the primary palm tree identification, achieving up to $85\%$ accuracy for palm trees.

This highlights a potential strategy: reducing false identifications in the primary object class by training the model on frequently misidentified objects. However, this demands extensive manual labeling, as seen with nearly $2,000$ okra plants in this case.

Subsequent experiments will utilize the green background and four object classes, given their demonstrated superiority in mAP@.5 outcomes.

\begin{table*}[htbp]
{\small
\begin{center}
\begin{tabularx}{\linewidth}{|l|X|l|l|}
\hline
\textbf{Abbr.} & \textbf{Obj. Classes} & \textbf{Cut out obj.} & \textbf{Labelled (25m)}\\
\hline
P. & palm trees & 44 & 187\\
P.O. & palm trees, okra plants & 44, 30 & 187, 2471\\
P.O.W.T. & palm trees, okra plants, weeds, tree trunks & 44, 30, 17, 24 & 187, 2471, 62, 193\\
\hline
\end{tabularx}
\end{center}
}
\caption[Set of object classes]{Set of object classes \label{tab:Object Class}}
\end{table*}

% %\ref{fig:Exp4}
% \begin{figure}[hpbt]
%  \centering
%   %%----start of first subfigure----
%   \subfloat[Test results with one, two and four classes]{
%    \label{fig:multipic:a} %% label for first subfigure
%    \includegraphics[width=\linewidth]{paper/images/Exp2.png}}\\
%  \caption[Image (a) shows the test result and image (b) a detect example]{Image (a) shows the test result and image (b) a detect example}
%  \label{fig:Exp4} %% label for entire figure
% \end{figure}

\subsection{Effect of Increasing Training Images on Accuracy}

This study investigates the potential correlation between increasing training images and the mAP@0.5 metric. The experiments conducted previously consistently used $300$ training and $120$ validation images. The most promising mAP@0.5 values emerged using the green BG and four object classes, which then served as our consistent benchmark, as detailed in Table \ref{tab:Increase Number}.

%\ref{tab:Increase Number}
\begin{center}
\begin{table}[htbp]
{\small
\begin{center}
\begin{tabular}{|l|l|l|}
\hline
\textbf{Training img} & \textbf{Validation img} & \textbf{$\varnothing$ Number of palms}\\
\hline
300 & 120 & T:6000 V:2400\\
600 & 240 & T:12000 V:4800 \\
\hline
\end{tabular}
\end{center}
}
\caption[Number of training and validation images]{Number of training and validation images \label{tab:Increase Number}}
\end{table}
\end{center}
 
Notably, doubling the training images to $600$ led to a drop in mAP@0.5 values, not surpassing the $0.8$ benchmark. This may suggest that using more images from the same source pool can lead to model overfitting, thus impairing its ability to generalize for unseen data. This is further evidenced by the rise in false positives across all classes.

Furthermore, enlarging the training dataset correspondingly extends the experiment duration, notably during the upload to Google Drive and subsequent Colab training.

%\ref{fig:Exp5.1}
\begin{figure}[htb]
 \centering
  %%----start of first subfigure----
  %\subfloat[Result of ...]{
   \label{fig:exp3:a} %% label for first subfigure
   \includegraphics[width=\linewidth]{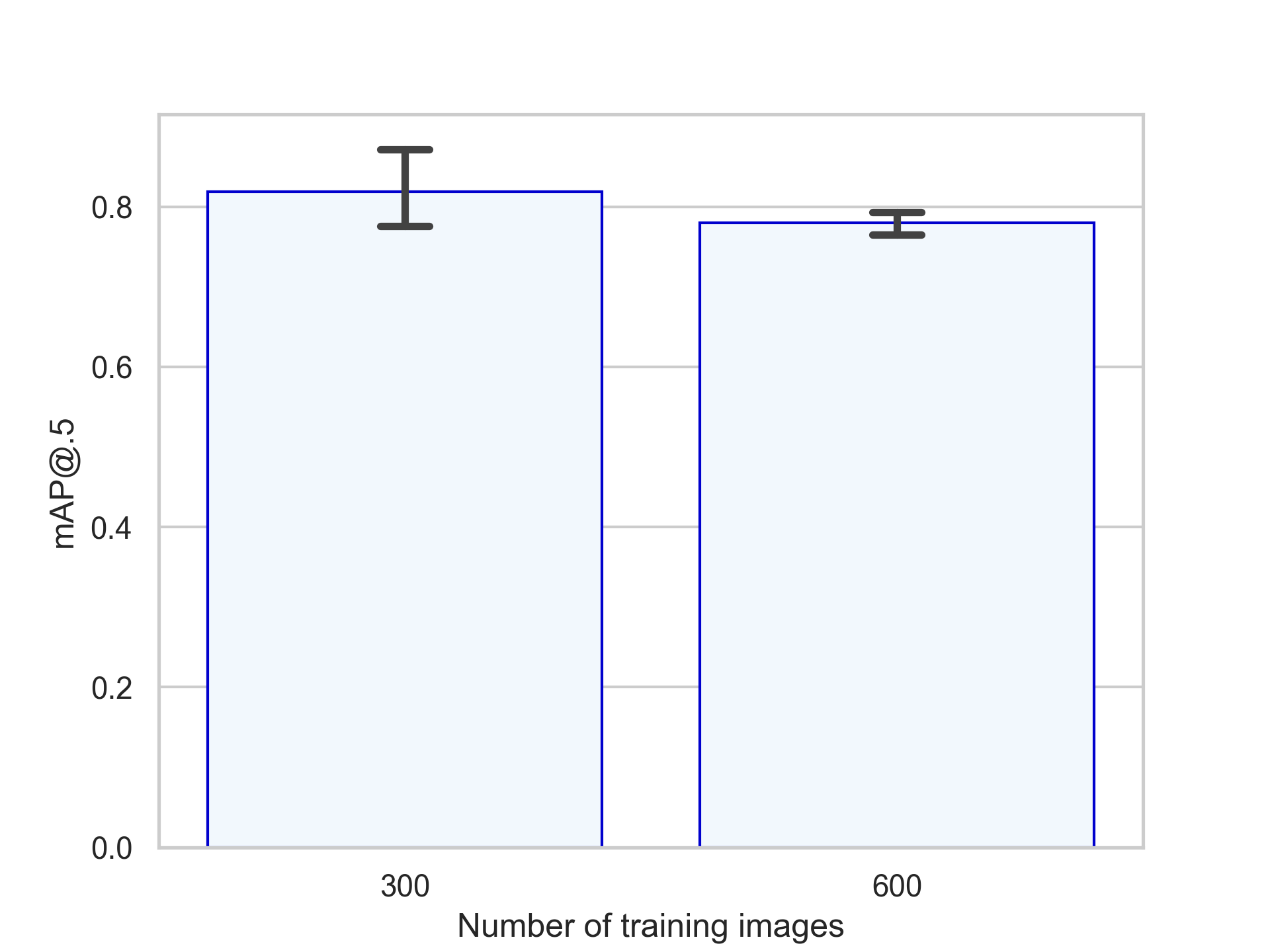}
 \caption[Increased number of training images with 4 classes and green BG]{Increased number of training images with 4 classes and green BG}
 \label{fig:Exp5.1} %% label for entire figure
\end{figure}

\subsection{Influence of Drone Altitude on Test Image Quality}

Until now, training and validation data primarily came from drone shots taken at approximately $10$m and $25$m above the ground. The test data was predominantly captured at $25$m. With the most consistent results using the green BG, four object classes, and 300 training images, this configuration became the baseline for this study.

With the goal of counting all coconut trees using a trained model, efficient drone photography becomes pivotal. A central question is determining the optimal drone altitude to capture the maximum number of trees without compromising image quality. While greater altitude captures more trees in one shot, the trees occupy fewer pixels, as detailed in Table~\ref{tab:Height}.

\begin{figure}[htb]
  \centering
  \includegraphics[width=\linewidth]{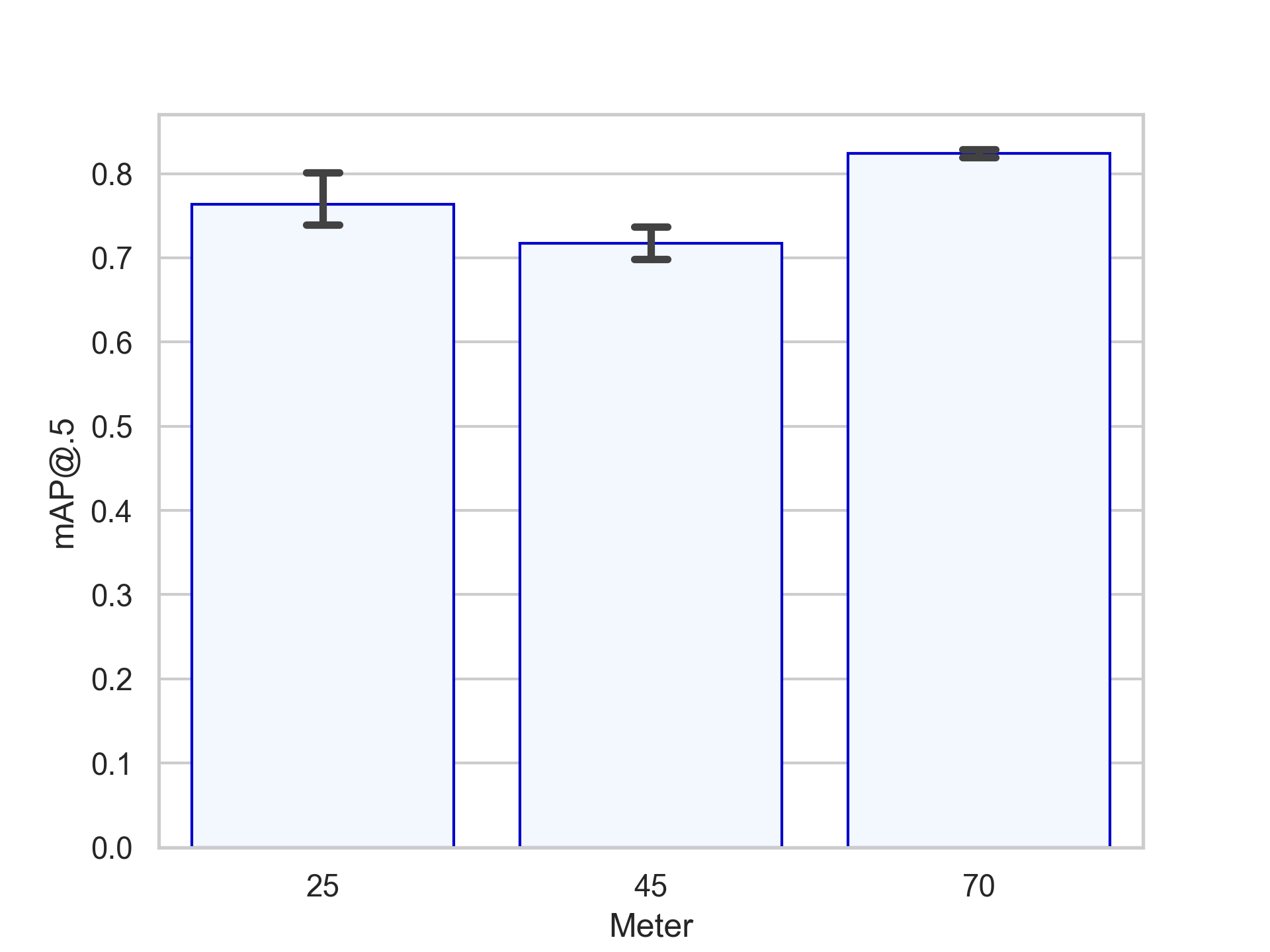}
  \caption[Height of test images]{Height of test images}\label{fig:Exp6}
\end{figure}

Results, visualized in Figure~\ref{fig:Exp6}, indicate that $70$m above ground provides the most promising test data. However, due to the scarcity of high-altitude footage, this conclusion serves only as an initial indicator. While $70$m seems to be the optimal altitude, this conclusion, based on only three test images with $66$ coconut trees, requires further validation. Furthermore, strategies must be developed to ensure comprehensive land coverage with drone shots, avoiding gaps or overlaps. Subsequent experiments will continue to rely on the $25$m data.

%\ref{tab:Height}
\begin{center}
\begin{table}[htbp]
{\small
\begin{center}
\begin{tabular}{|l|l|l|}
\hline
\textbf{Height of test images} & \textbf{Image count} & \textbf{Total palms} \\
\hline
25m & 38 & 187 \\
45m & 12 & 126\\
70m & 3 & 66\\
\hline
\end{tabular}
\end{center}
}
\caption[Number of test images and palms in total]{Number of test images and palms in total \label{tab:Height}}
\end{table}
\end{center}

%\ref{fig:Exp6}.

\subsection{Number of Palm Trees on Training Images}

\begin{figure*}[hpbt]
 \centering
  %%----start of first subfigure----
  \subfloat[Number of palm trees]{
   \label{fig:Exp7} %% label for first subfigure
   \includegraphics[width=0.46\linewidth]{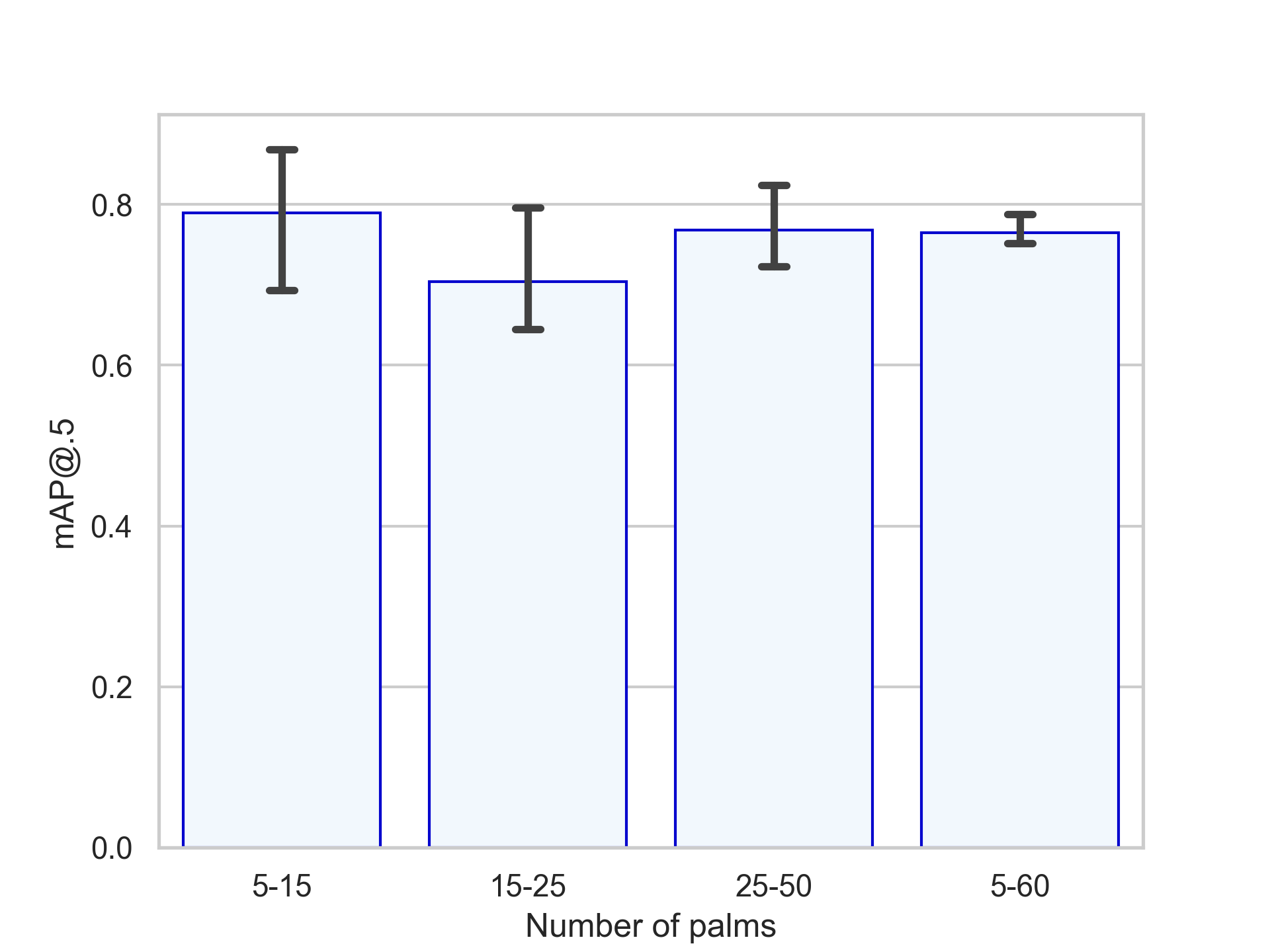}}
  \hspace{0.01\linewidth}
  %%----start of second subfigure----
  \subfloat[Range of palm trees]{
   \label{fig:Exp8} %% label for second subfigure
   \includegraphics[width=0.46\linewidth]{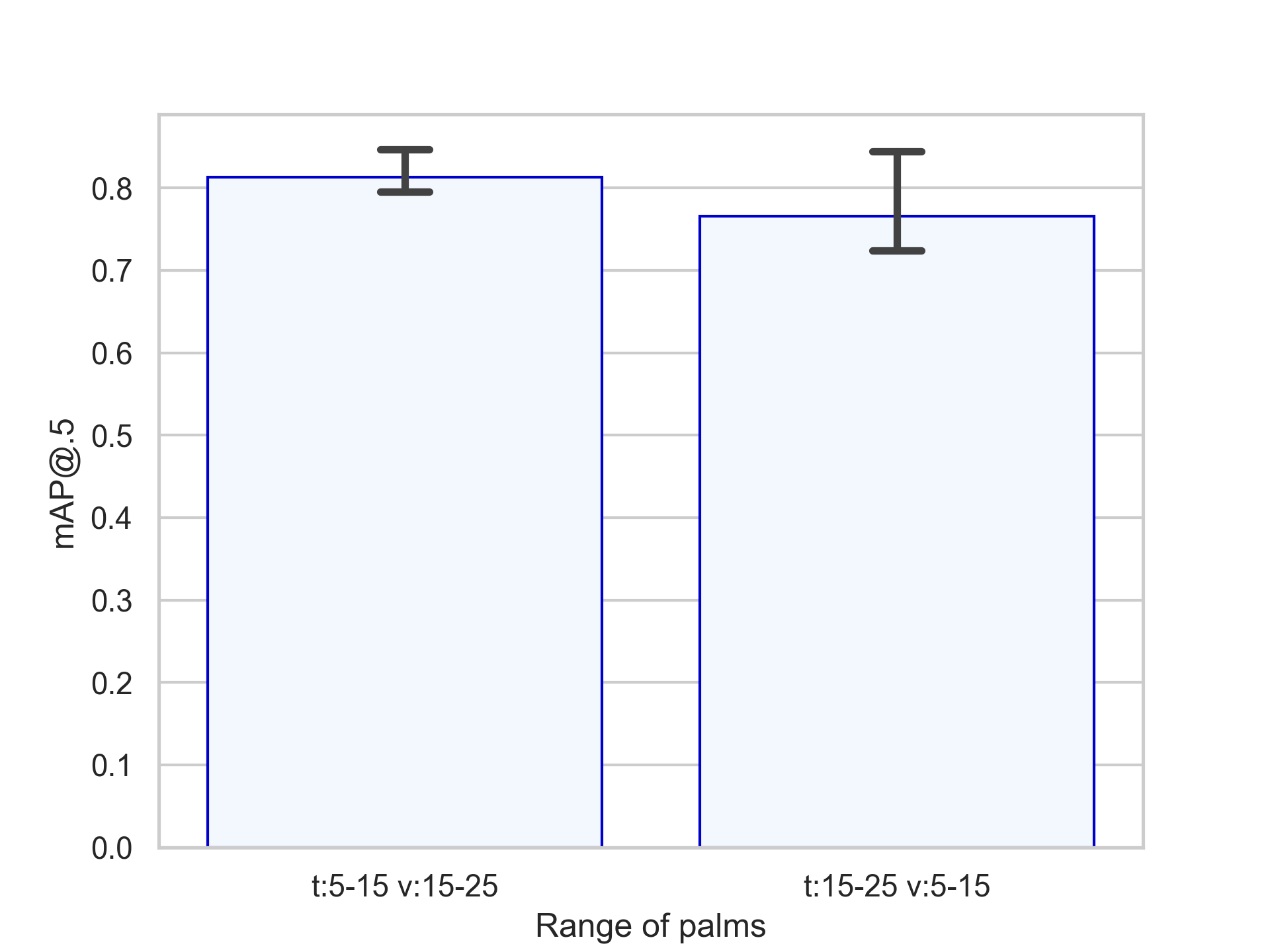}}\\[0pt] % horizontal break
 \caption{Numbers and range of palm trees in training and validation}
\end{figure*}

We investigated the impact of varying the number of palm trees placed on backgrounds during training and validation data generation. Prior experiments employed a random placement of 15 to 25 palm trees per image. Table \ref{tab:Range} enumerates the chosen ranges, with results visualized in Figure \ref{fig:Exp7}. It's notable that our model incorporated a green BG, identified the coconut palm tree as its single object class, and was trained using 300 synthetic images.

A range of 5 to 15 palm trees yielded the highest mAP@.5 scores. Given the average palm tree count of 13 in the test drone footage, the model possibly aligned better with test images having 2 to 14 palms. This suggests that training and validation data with similar palm tree ranges potentially improves model assumptions on test data. On the contrary, the 15 to 25 range resulted in the lowest mAP@.5, possibly due to the absence of this tree range in test images, suggesting the model's assumptions based on training data might misalign. The 25 to 50 range saw mAP.5 improvements, likely reflecting the tree counts in test data.

This theory warrants further exploration to better understand model internalizations and their effects. A subsequent experiment delves deeper into this.

% \begin{figure}[htb]
%   \centering
%   \includegraphics[width=\linewidth]{paper/images/Exp5.png}
%   \caption[Number of palm trees in training and validation data]{Number of palm trees in training and validation data}\label{fig:Exp7}
% \end{figure}

% \begin{figure}[hbt]
%  \centering
%   %%----start of first subfigure----
%   \subfloat[Number of palm trees in training and validation data]{Number of palm trees in training and validation data]{
%    \includegraphics[width=0.46\linewidth]{paper/images/Exp5.png}\label{fig:Exp7}}
%    \hfill
%   %%----start of second subfigure----
%   \subfloat[Different number of palm trees in training and validation]{
%    \label{fig:Exp8} %% label for second subfigure
%    \includegraphics[width=0.46\linewidth]{paper/images/Exp6.png}}
%  \caption{TODO}
% \end{figure}

%\ref{tab:Range}
\begin{center}
\begin{table}[htbp]
{\small
\begin{center}
\begin{tabularx}{\linewidth}{|X|X|}
\hline
\textbf{Palms per image} & \textbf{$\varnothing$ Number of palms} \\
\hline
5 to 15 & 3000 \\
15 to 25 & 6000\\
25 to 50 & 11250\\
5 to 60 & 9750\\
\hline
\end{tabularx}
\end{center}
}
\caption[Number of palm trees in training and validation images]{Number of palm trees in training and validation images \label{tab:Range}}
\end{table}
\end{center}

\subsection{Different Palm Count for Training and Validation}

Up to now the number range of palms in training and validation data was equal. In the previous experiment a theory was set up that there 
might be some internalisation in the model that the number of palm trees in validation and test are similar to those in training. In this section the range used for training and validation data differs. Figure \ref{fig:Exp8} depicts 3 variants, the first two variants with different ranges and the last variant with same ranges for comparison. This is summarized in table \ref{tab:RangeTV}. The experiment explores if there is an improvement in the mAP.5 values by having different training and validation ranges utilised for the training of the model. It was trained with a green BG, one object class (coconut palm trees) and 300 synthetic images.

The first variant shows slightly higher mAP@.5 values, which makes it one of the best results so far. The second variant depicts lower mAP@.5 values. The third variant show the result of the previous section.

One possible explanation for the first variant could be that by training the model with different training and validation ranges, it can handle the variety of palm trees in each test image better. That means the detection and classification of palm trees are more accurate. Another reason could be that the total number of palms and their split in training and validation is more important than the values per image.  

% \begin{figure}[htb]
%   \centering
%   \includegraphics[width=\linewidth]{paper/images/Exp6.png}\\
%   \caption[Different number of palm trees in training and validation]{Different number of palm trees in training and validation}\label{fig:Exp8}
% \end{figure}

%\ref{tab:RangeTV}
\begin{center}
\begin{table}[htbp]
{\small
\begin{center}
\begin{tabularx}{\linewidth}{|X|l|l|}
\hline
\textbf{Palms per image} & \textbf{$\varnothing$ Palms (train)}  & \textbf{$\varnothing$ Palms (val)}\\
\hline
t:5-15 v:15-25 & 3000 & 2400 \\
t:15-25 v:5-15 & 6000 & 1200\\
t:5-15 v:5-15 & 3000 & 1200\\
\hline
\end{tabularx}
\end{center}
}
\caption[Number of palm trees in training and validation images]{Number of palm trees in training and validation images \label{tab:RangeTV}}
\end{table}
\end{center}

\subsection{Freezing Layers}

\begin{figure}[hpbt]
  \centering
  \includegraphics[width=\linewidth]{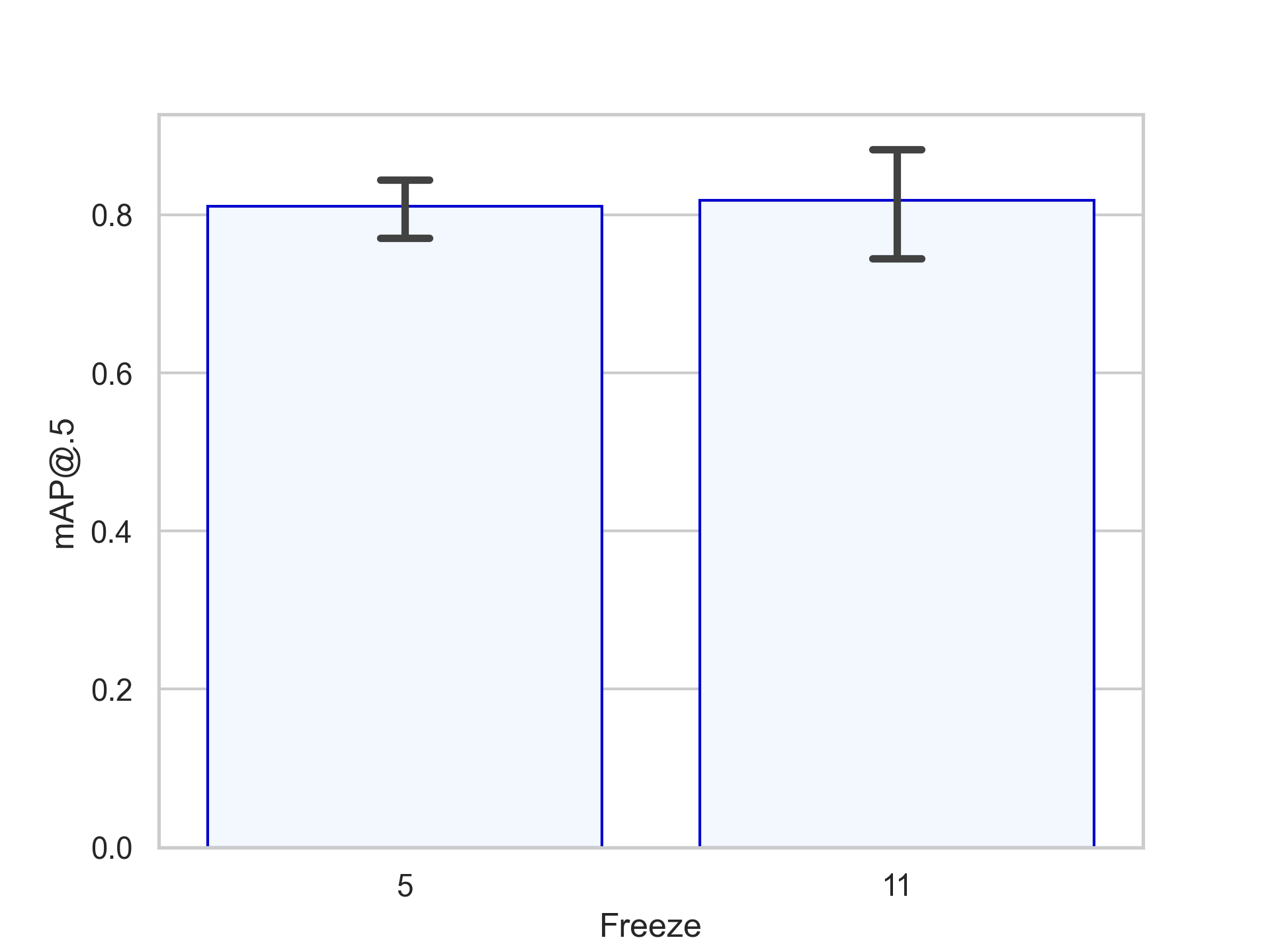}\\
  \caption[Freezing initial layers from 0-4 and 0-10]{Freezing initial layers from 0-4 and 0-10}\label{fig:Exp9}
\end{figure}

Yolov7, as detailed in section 2.4, boasts various layers with its initial weights trained on the Coco dataset. When we freeze a layer, we prevent its weights from updating during training. Typically, the architecture's initial layers capture basic features like edges. The backbone of YOLOv7 encompasses 50 layers. In this experiment, we adjusted the \textit{freeze} hyperparameter for fine-tuning. Figure \ref{fig:Exp9} presents outcomes when fixing the first 5 and 11 layers. We trained using a green BG, four object classes, and 300 images over 5 repetitions. The mAP@.5 values remained consistent across both frozen and non-frozen variants, suggesting that the \textit{freeze} hyperparameter might not provide added benefits. Detection results revealed a high true positive rate: of the 187 palm trees labeled in ground truth, 199 were detected with minimal false positives.

%\ref{fig:Exp9}

\subsection{Agricultural Implications}

Through the conducted experiments, we successfully elevated mAP@0.5 from an initial 0.65 to an average of 0.88. For agricultural planning, such as procuring fertilizers and protective nets, this accuracy proves sufficient. Table \ref{tab:summery} consolidates the baseline and optimal variants from each experiment. Notably, in the final \textit{freeze} experiment, out of 187 labeled palm trees in the ground truth, 199 were accurately detected with minimal false positives. While all experiments include interpretations, further testing is essential for validation, given the inherent variability in mAP.5 results even under consistent parameters. Enhancing model transparency and clarity can also benefit from advances in explainable AI.

%\ref{tab:summery}

\begin{table*}[htb]
{\small
\begin{center}
\begin{tabularx}{\linewidth}{|X|X|X|X|X|l|l|}
\hline
\textbf{BG} & \textbf{Classes} & \textbf{Train Img} & \textbf{Palms per Img} &  \textbf{Freeze} & \textbf{Min mAP} & \textbf{Max mAP} \\
\hline
Red & Palms & 300 & 15-25 & none & 0.63 & 0.69 \\
Green & Palms & 300 & 15-25 & none & 0.73 & 0.80 \\
Green & Palms & 300 & T:5-15 V:15-25 & none & 0.79 & 0.84 \\
Green & Palms & 300 & 5-15 & none. & 0.69 & 0.86\\
Green & P.O.W.T. & 300 & 15-25 & none & 0.77 & 0.87\\
Green & P.O.W.T. & 300 & 5-15 & 0-10 & 0.75 & 0.88\\
\hline
\end{tabularx}
\end{center}
}
\caption[Comparison of best variants of selected experiments]{Comparison of best variants of selected experiments \label{tab:summery}}
\end{table*}

\section{CONCLUSION}

This research tackled the challenge of counting coconut palm trees in drone imagery using deep learning. Among various real-time object detectors, YOLOv7 from the YOLO family emerged as the preferred choice. Given the limited availability of drone footage, we strategically generated synthetic images to train and validate our model. As a result, we witnessed a significant enhancement in the mAP@.5 value, elevating it from 0.65 to 0.88. We varied input parameters and fine-tuned hyperparameters, finding that the generation of synthetic images, particularly with stable diffusion backgrounds, was beneficial. Our best model detected 199 palm trees out of 187 labelled in the test data, with minimal false positives.

For comprehensive results, systematic drone capture of land remains essential. Additional fine-tuning and experiments can further optimize the model. Transitioning this methodology into a scalable product holds potential for aiding nearby farms in yield estimation and strategic planning. Future research could also focus on assessing the health of coconut palms. By integrating traditional farming with advanced techniques, we aimed to transform manual surveys into a semi-automated, cloud-based solution. This approach is not only cost-effective but also reduces time, labor, and errors. Bridging agriculture with technology, especially through drones and deep learning, unveils a horizon of promising opportunities.

\section*{\uppercase{Acknowledgements}}
This paper was partially funded by the German Federal Ministry of Education and Research through the funding program "quantum technologies - from basic research to market" (contract number: 13N16196). Furthermore, this paper was also partially funded by the German Federal Ministry for Economic Affairs and Climate Action through the funding program "Quantum Computing -- Applications for the industry" (contract number: 01MQ22008A).

\bibliographystyle{apalike}
{\small
\bibliography{references}}

% \section*{\uppercase{Appendix}}

% If any, the appendix should appear directly after the
% references without numbering, and not on a new page. To do so please use the following command:
% \textit{$\backslash$section*\{APPENDIX\}}

\end{document}